\documentclass{bmvc2k}
\usepackage{indentfirst}
\usepackage{multirow}
\usepackage{amsmath,amssymb}
\title{Semi-Supervised Cross-Modal Salient Object Detection with U-Structure Networks}

\addauthor{Yunqing Bao}{yunqing.bao@mbzuai.ac.ae}{1}
\addauthor{Hang Dai}{hang.dai@mbzuai.ac.ae}{2}
\addauthor{Abdulmotaleb Elsaddik}{a.elsaddik@mbzuai.ac.ae}{3}

\addinstitution{
 Yunqing Bao \\
 MBZUAI, Abu Dhabi, UAE \\
 Hang Dai \\
 MBZUAI, Abu Dhabi, UAE \\
 Abdulmotaleb Elsaddik \\
 MBZUAI, Abu Dhabi, UAE
}

\runninghead{Student, Prof}{Yunqing Bao, Hang Dai, Abdulmotaleb Elsaddik}

\begin{document}

\maketitle

\begin{abstract}
   \textit{Salient Object Detection (SOD) is a popular and important topic aimed at precise detection and segmentation of the interesting regions in the images. We integrate the linguistic information into the vision-based U-Structure networks designed for salient object detection tasks. The experiments are based on the newly created DUTS Cross Modal (DUTS-CM) dataset, which contains both visual and linguistic labels. We propose a new module called efficient Cross-Modal Self-Attention (eCMSA) to combine visual and linguistic features and improve the performance of the original U-structure networks. Meanwhile, to reduce the heavy burden of labeling, we employ a semi-supervised learning method by training an image caption model based on the DUTS-CM dataset, which can automatically label other datasets like DUT-OMRON and HKU-IS. The comprehensive experiments show that the performance of SOD can be improved with the natural language input and is competitive compared with other SOD methods.}
\end{abstract}

%%%%%%%%% BODY TEXT
\section{Introduction}
Salient Object Detection (SOD) is a topic about making computer to mimic the pre-attentive stage of human visual system to segment the highly interesting regions that could draw more attention in the images \cite{gupta2020salient}. There are lots of methods proposed in the domain of Salient Object Detection, including the conventional methods and the deep learning based methods. And the success of the deep convolutional neural network (CNN) creates a great impact on the domain of SOD. Many researchers proposed deep learning based methods for SOD, and those methods could generate better performance than conventional methods \cite{klein2011center}.

Among the deep learning based methods, U-Structure methods are quite suitable for the SOD tasks, as they could combine low-level and high-level features efficiently. The classic U-Structure methods are U-Net \cite{ronneberger2015u} and U$^2$Net \cite{qin2020u2}. All of those methods focus on the vision information, while in different modalities other than vision may also be able to draw attention to detect and segment the salient image regions. The two most important modalities are vision and language. Segmentation tasks based on vision and language is called Referring Image Segmentation, whose goal is to segment the image regions or the objects referred by the language expression in the images. It gives an inspiration that adding linguistic information to the vision-based SOD structures could help to improve the accuracy. 

To prove the inspiration, we propose a new dataset named DUTS-CM, which is based on DUTS \cite{wang2017learning}. In this dataset, we label each image in this dataset with a pixel-level annotated binary mask image, but also captions to describe the contents of the images. This dataset is also used for training an image caption model to make the labelling of other datasets easier. And to better integrate the features from two modalities, a module named efficient Cross-Modal Self-Attention (eCMSA) is created to conflate the features extracted from vision and text to further improve the performance of the U-Structure networks.

In summary, our work generally has following contributions:
(1) Labelling three datasets (DUTS, DUT-OMRON, HKU-IS) with linguistic captions. DUTS is completely manually labeled, while the image caption model labels the other two datasets.
(2) We propose a cross-modal self attention module to fuse visual and textual representations efficiently. This module could help to improve the performance of some U-structure methods and can achieve quite competitive metrics on all the three datasets.
(3) Except for the two major contributions mentioned above, we also release an English version of the visual-linguistic labeling tool for the benefits of the vision-language community.  

\section{Related Work}
\subsection{Salient Object Detection}
The main proposal of salient object detection is to detect or segment the most salient part of objects or pixels in the natural scenes. And a detailed survey is provided in \cite{borji2019salient}. 

There are some conventional methods, such as \cite{srivatsa2015salient}, \cite{liu2010learning} and \cite{zhang2015minimum}. Those models acquire the saliency maps mainly by extracting the hand-crafted features. But this method always suffering from its unsatisfying low generalization ability for challenging situations. Then the surge in popularity of deep neural networks after year 2012 makes deep learning based methods the mainstream direction nowadays, which can be divided into three groups, classic convolutional methods, Fully Convolutional Network(FCN) \cite{long2015fully} related methods and others. 

\textbf{Classic convolutional methods.}
This kind of methods are usually related to the CNN architecture. Wang \cite{wang2015deep} combined both the global and local features to segment the salient parts of images. Kim \cite{kim2016shape} created a two-branch CNN network which can combine global and local context and estimate the shape of salient objects. Region proposals are generated by the selective search method \cite{uijlings2013selective}, and saliency maps are refined through hierarchical segmentation. Li and Yu \cite{Li_2015_CVPR} extracted multi-scale features from images and enhance the spatial coherence by a superpixel based saliency refinement method.

\textbf{FCN based method.}
Except for classic convolutional methods, some deep learning based methods are related with FCN, which considers pixel-level information. Zhang \cite{zhang2017learning} proposed a simplified method based on FCN which can tackle with the issues of checkerboard artifact through a new up-sampling method and learn deep uncertain convolutional features(UCF). Tang \cite{tang2016saliency} proposed a VGGNet \cite{simonyan2014very} based network which fuses pixel-level and region-level saliencies. Hu tackled blurred saliency and inaccurate boundaries issues by building a deep Level Set network \cite{osher1988fronts} and a Guided Superpixel Filtering layer.

\textbf{Other methods.}
Also, there are some other methods based on deep recurrent or attention structures. Liu \cite{liu2018picanet} proposed PiCANet, a network generating attention for context regions of pixels. Then the authors incorporated this network with the U-Net structure to detect and segment the salient region of pixels. Wang \cite{wang2019salient} created the PAGE-Net by adding the pyramid attention structure and the salient edge detection module to the typical bottom-up/top-down structure. This kind of design could enhance the representation ability of the network and achieve better edge-preserving segmentation.

\subsection{Referring image segmentation and attention}
\textbf{Referring image segmentation}
The proposal of referring image segmentation is to find the referent in images and precisely generate a segmentation mask, which is different from referring image localization. As an early attempt, \cite{hu2016segmentation} firstly proposed the problem of referring image segmentation and the typical LSTM-CNN model \cite{6795963} is created to approach this problem. In this architecture, CNN and LSTM are used to extract visual and linguistic features separately. Then they are concatenated to predict the segmentation mask. \cite{margffoy2018dynamic} proposed a novel network called Dynamic Multimodal Network (DMN). The author replaced the standard LSTM with Simple Recurrent Units (SRUs) \cite{lei2018training} to improve the speed and built a Synthesis Module(SM) to fuse the visual and textual features in a better way. To tackle the issue of the lack of multi-scale semantics, Li \cite{li2018referring} proposed a Recurrent Refinement Network (RRN). It firstly extracts visual features and linguistic features through CNN and LSTM, then concatenates the two kinds of features, and processes them hierarchically and recurrently through convolutional LSTM \cite{shi2015convolutional}.

\textbf{Attention mechanism}
Attention has become a powerful tool in many areas, such as computer vision, natural language processing \cite{bahdanau2014neural,carion2020end}. And in the area of vision and language, attention is also widely used as its effectiveness for multi-modalities. Our work is also a kind of vision and language task relating to attention. Among the attention-based architectures, self-attention \cite{vaswani2017attention} is a very popular and effective one recently proposed, whose major function is to relate different parts of a single sequence and correspondingly generate a representation of this sequence. Based on self-attention, a novel architecture named Transformer is proposed to approach problems not only like machine translation but many other tasks like text classification \cite{sun2019fine}, visual question answering, visual commonsense reasoning \cite{li2019visualbert}.

\section{Dataset}
\subsection{DUTS-CM}
The new dataset, DUTS Cross Modal(DUTS-CM), is completely based on the DUTS dataset. DUTS is a saliency detection dataset, it contains 10553 training images and 5019 test images. The training images are from the ImageNet dataset, and the test images are from both the ImageNet and the SUN datasets. And there are more challenging scenarios in test images than training images, which helps to check the robustness of the algorithm.

For DUTS-CM, the train/test datasets are split in the same way as DUTS. Every image in the dataset is labeled with one or two simple sentences to describe the content of the image. The sentences mainly describe the name and color of the objects, sometimes will also describe the color of the background and the spatial relationship between different objects or between the objects and the background.  And all the images in DUTS-CM are annotated consistently by human labour, and almost all the annotations are created in the same pattern. Figure 1 shows how the images are manually labeled through an interface. 

\begin{figure}[!h]
\centering\includegraphics[scale = 0.30]{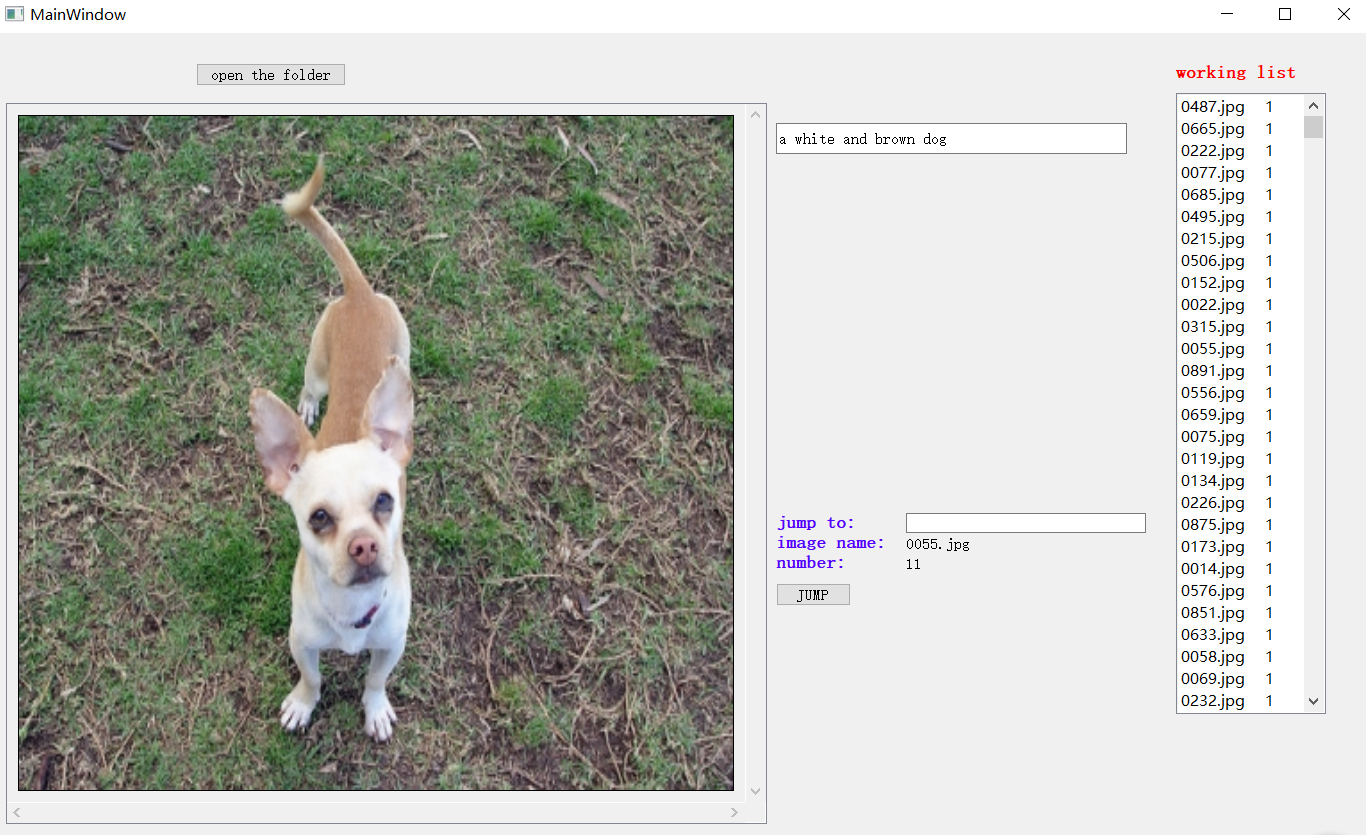}
\caption{The interface of the labeling tool, which is mainly based on this image caption labeling tool method \href{https://github.com/Erichen911/ImageCaptionLabelMaster}{https://github.com/Erichen911/ImageCaptionLabelMaster}.}
\end{figure}

\textbf{Labeling rules}
The labelling rules change with different scenarios, which could be roughly divided into simple and complex scenarios.

\begin{figure}[htbp]
\centering
\subfigure[]
{
    \begin{minipage}[b]{.45\linewidth}
        \includegraphics[width=60mm]{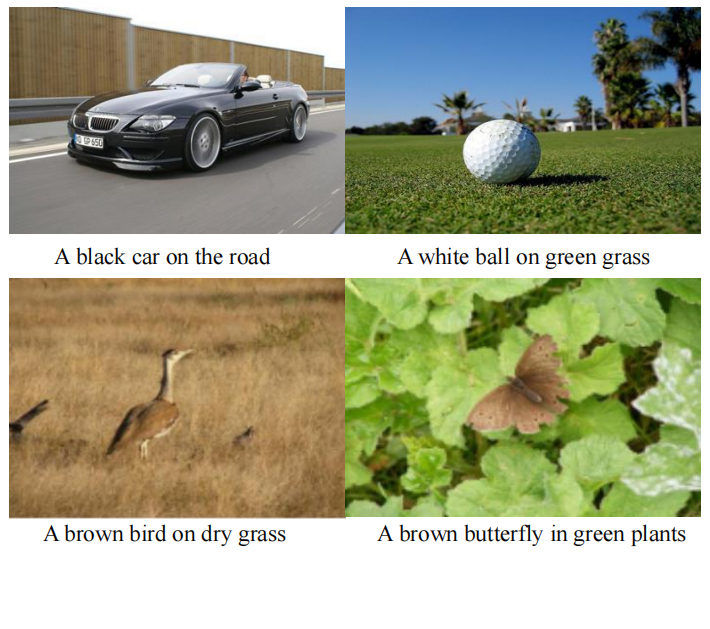}
    \end{minipage}%
}
\subfigure[]
{
 	\begin{minipage}[b]{.45\linewidth}
        \includegraphics[width=60mm]{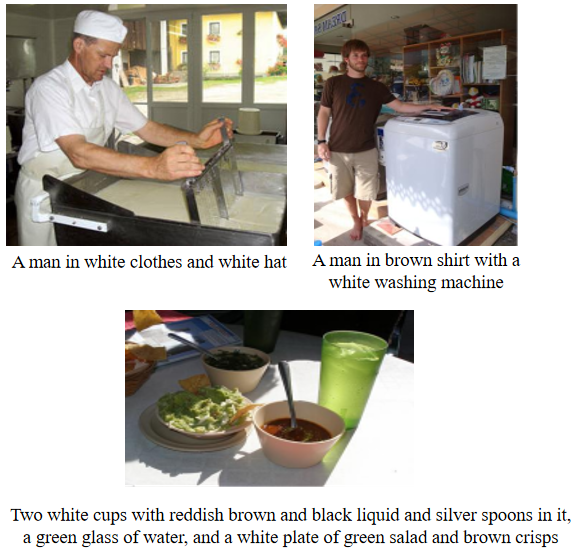}
    \end{minipage}%
}
\caption{(a): simple scenarios, (b): complex scenarios}
\end{figure}

For simple scenarios, as shown in figure 2(a), only the colors and names or other obvious features of the single salient objects will be roughly described. If the backgrounds of the images are relatively monotonous and simple, the colors, names or other features of backgrounds will also be described. For those simple scenarios, usually one or two short sentences will be enough for a description. 

And for complex scenarios, as shown in figure 2(b), there are multiple objects in the images, some of them may overlap with each other, or appear with mixed colors. Under these circumstances, only the most salient objects are described with their colors and object classes, and also other consistent features. For complex scenarios, we usually need more and longer sentences to describe the salient part of the images.

The goal of SOD is to help machines learn how to distinguish the “important” parts of images like a human brains. But a human brains always follow general rather than strict rules. For example, it will roughly divide the colors into black, white, yellow, and so on, but not into some fine grained colors like magenta and absinthe green. Thus our dataset is also labeled with general descriptions.

\begin{figure}[!h]
\centering
\subfigure{\includegraphics[width = 60mm]{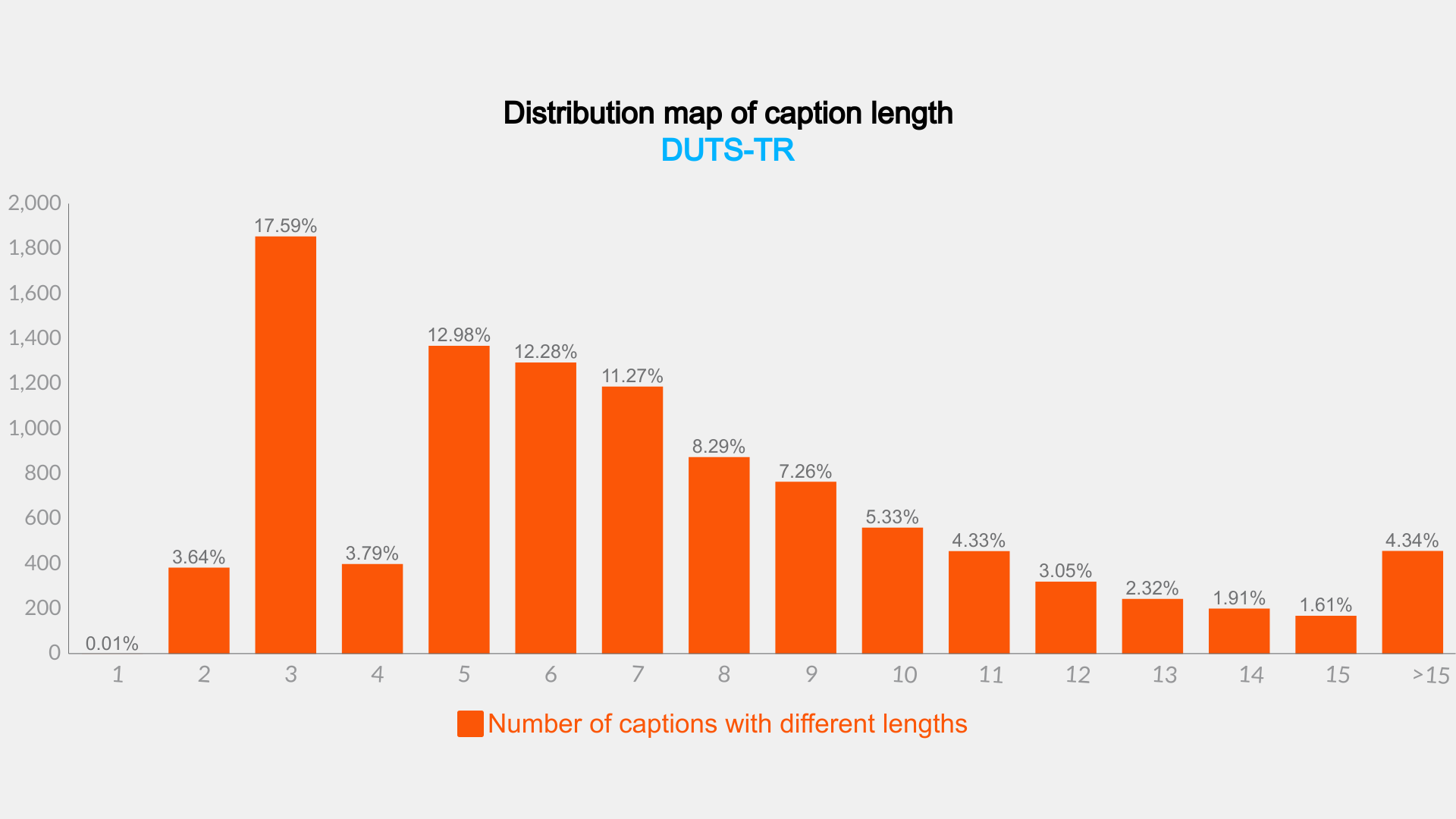}}
\subfigure{\includegraphics[width = 60mm]{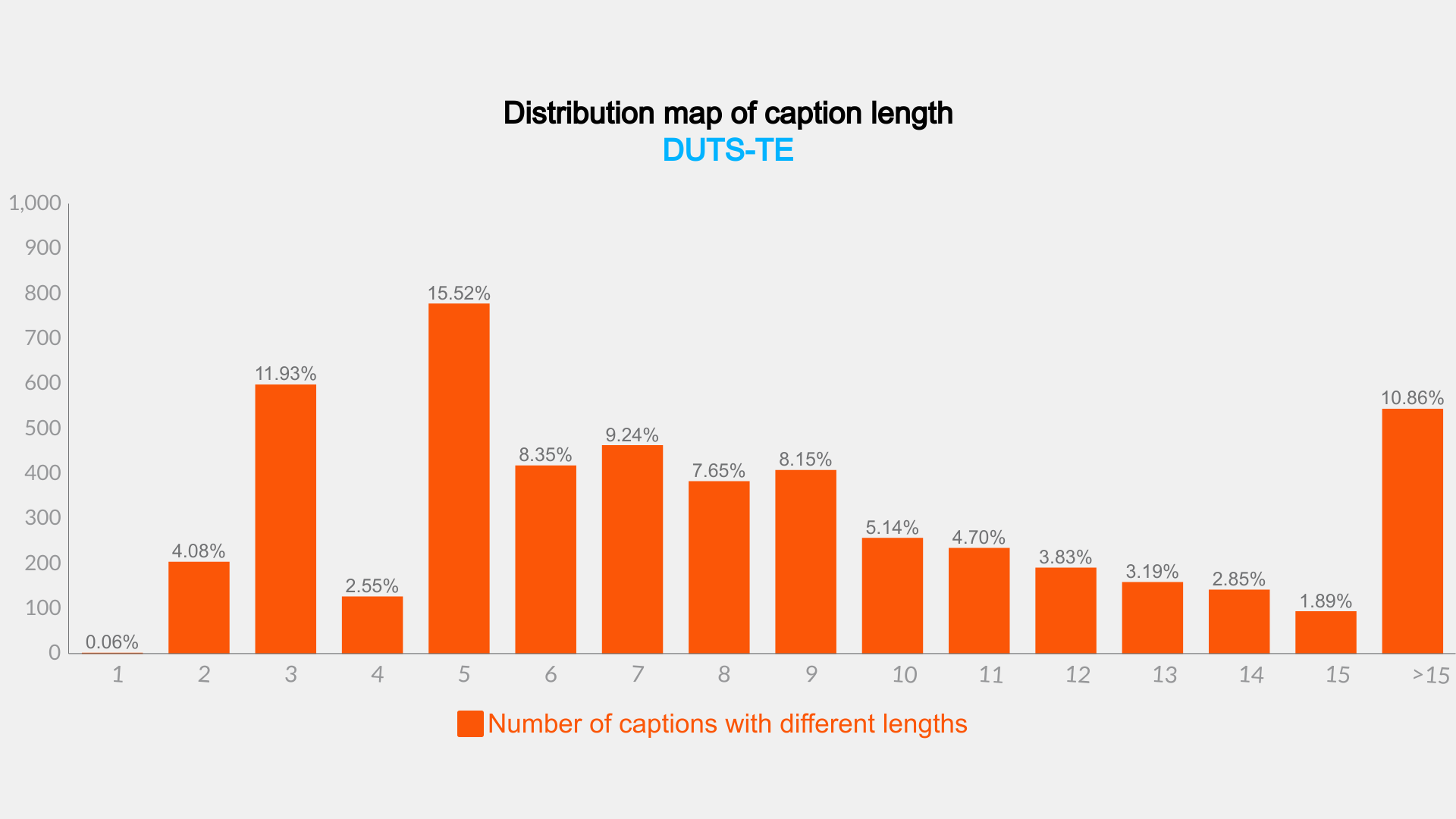}}\vspace{0pt}
\centering
\subfigure{\includegraphics[width = 60mm]{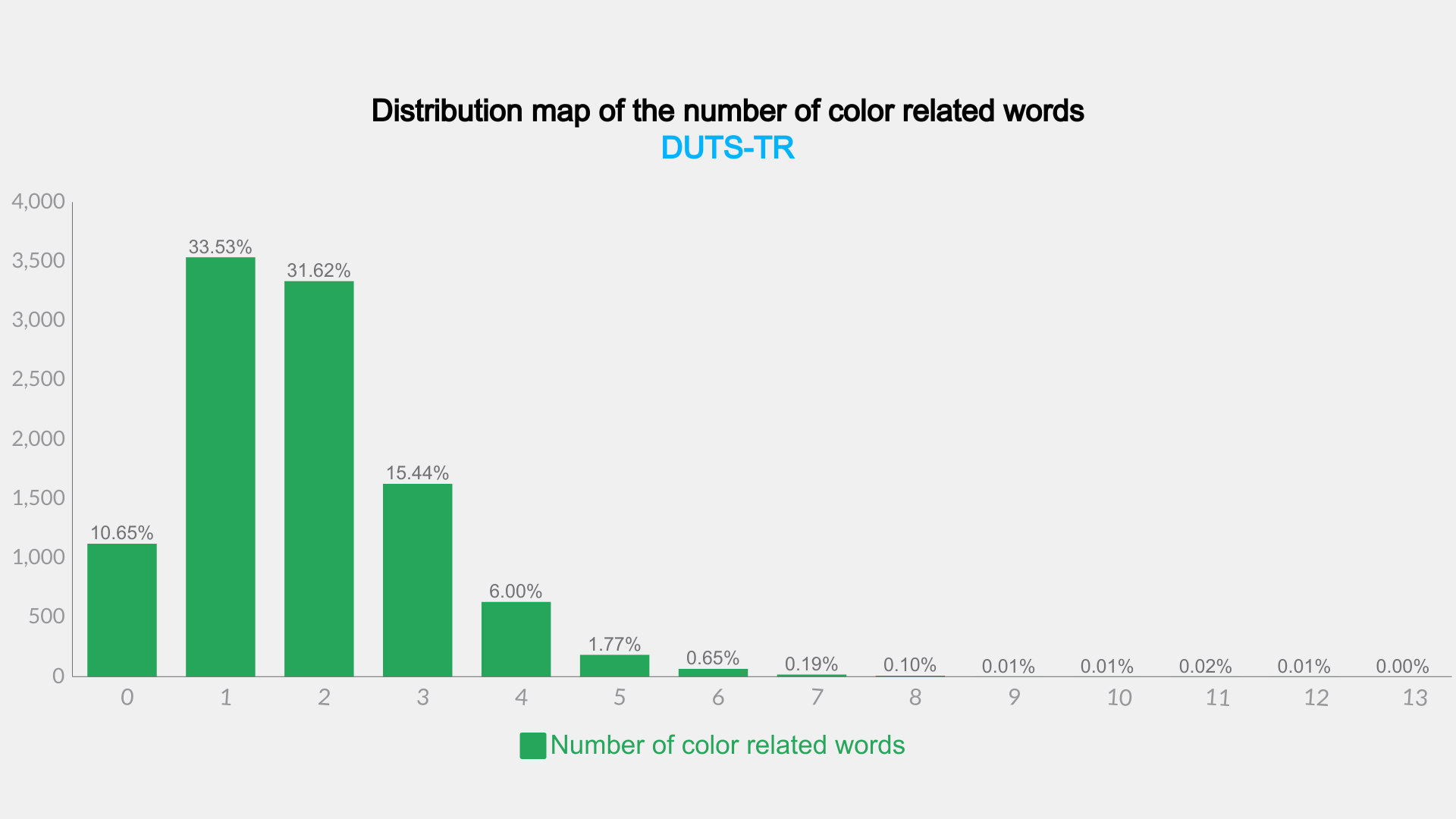}}
\subfigure{\includegraphics[width = 60mm]{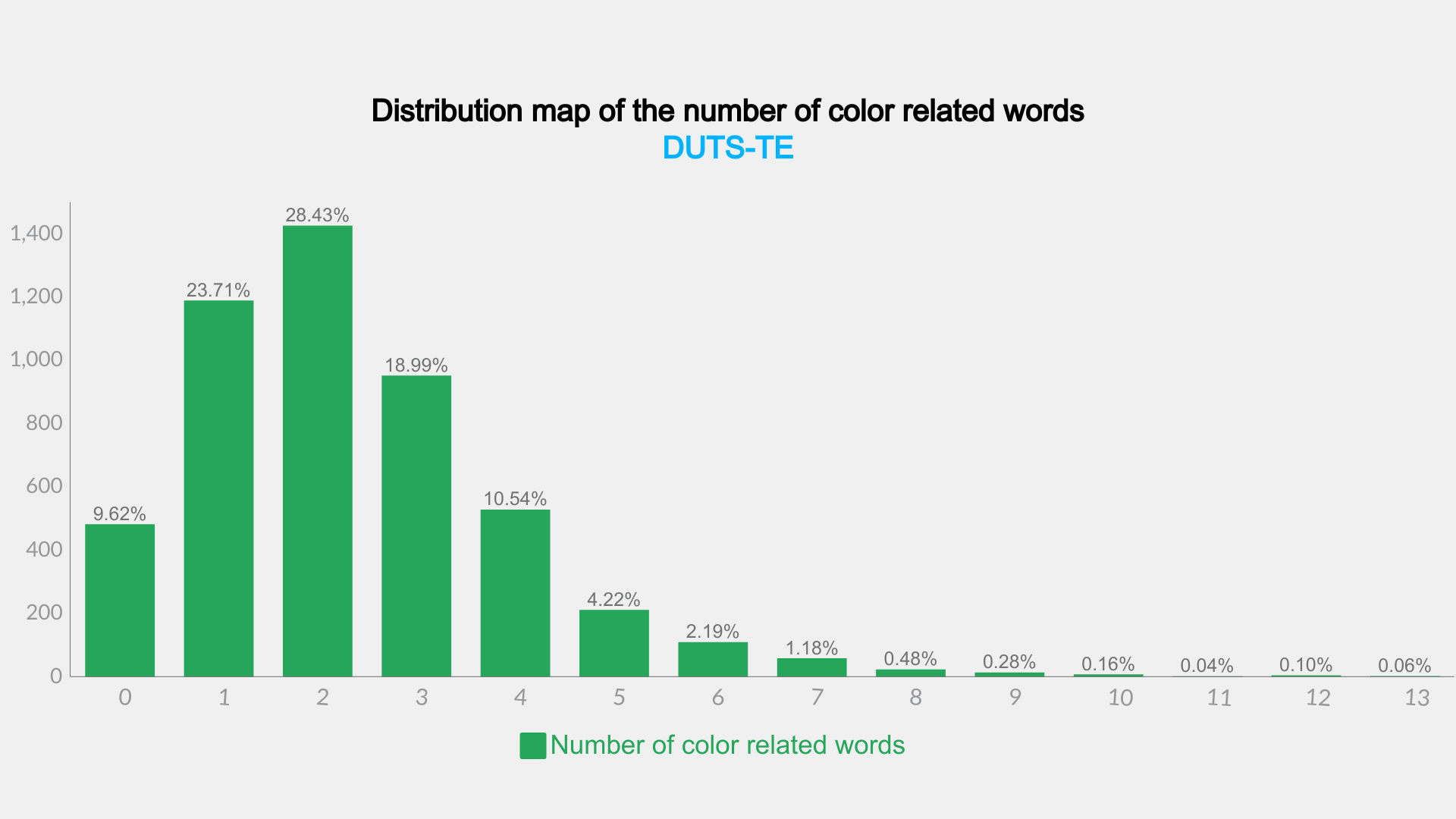}}
\caption{Statistical distribution of the caption length (how many words included in each caption) and color related words (how many color related words included in each caption). For all the four figures, the horizontal axis represents the length of the caption, and the vertical axis represents the number of captions in the dataset.}
\end{figure}

\textbf{Statistical properties of DUTS-CM}
From figure 3, we could see the distribution map of the caption (one caption for each image) length in the DUTS-CM dataset. Most of the captions in both DUTS-TR and DUTS-TE datasets are short ones which only contain 2 - 5 words. The average length of the captions in DUTS-TR is 7.26 words, while the average caption length of DUTS-TE is 8.71 words. Also, from the distribution map, we find the longer captions in DUTS-TE are more common to appear. Thus, maybe we could say that the secenarios of DUTS-TE are a little more complex than DUTS-TR. And we also find that the distribution of the caption length of DUTS-TE is similar to DUTS-TR. Besides, we added the color related words to the captions to better describe the scenarios. Most of the captions have at least one color related word. And about 85$\%$ of the captions have less than 5 color related words. Meanwhile, we could also find that the distribution of color related words in DUTS-TE is similar to DUTS-TR.

\subsection{DUT-OMRON-CM and HKU-IS-CM}
Except for DUTS, DUT-OMRON \cite{jiang2011automatic} and HKU-IS \cite{Li_2015_CVPR} are also datasets for the Salient Object Detection task with high quality and sufficient number of images. DUT-OMRON has 5168 well-labeled images, which is a quite challenging dataset for the SOD task. While HKU-IS has 4447 images, with simpler scenarios compared with DUTS and DUT-OMRON, meaning it is easier for the algorithms to achieve a quite high performance.

And the images of these two datasets are also labeled with descriptive sentences, not by human resources, but by an image caption model, trained on the DUTS-CM dataset. The link of the model is \href{https://github.com/saahiluppal/catr}{https://github.com/saahiluppal/catr}. This image caption model is generally based on the DEtection TRansformer(DETR) algorithm \cite{carion2020end}. Only the prediction module of DETR is replaced with the multilayer perceptron (MLP) to predict the captions as output. This also makes our entire training process semi-supervised. Then we have DUT-OMRON Cross Modal (DUT-OMRON-CM) and HKU-IS Cross Modal (HKU-IS-CM). Their captions are labelled by the image caption algorithm, which means there will inevitably be some mistakes, so we do not discuss their statistical properties.

\section{Method}
\subsection{BERT}
Firstly, the embedding of the sentence is obtained from BERT \cite{devlin2018bert}. BERT is a transformer network, which can be utilized for multiple natural language processing tasks, such as word embedding, named entity recognition and machine translation. In fact, BERT is efficient and probable to generate outstanding performances and state-of-the-art metrics on natural language processing tasks. And BERT is a pre-trained model, which means it could be easily transferred to a downstream task by fine-tuning. For this project, the [CLS] token of the last hidden state of BERT, which is originally used for classification tasks, is taken as the representation of the entire sentence. This representation is a vector with a length of 768.

\subsection{efficient Cross-Modal Self-Attention}
 For our work, a self-attention structure is similar to the Cross-Modal Self-Attention

 (CMSA) \cite{ye2019cross} structure is created to merge the features from two modalities. CMSA, whose structure is shown in figure 4(a), is a module in which the long-range dependencies could be obtained and different spatial attentions can also be captured from the conflated feature maps. From a CNN layer, a visual feature map can be represented as $F_1\in R^{H\times W\times C}$. And $F_2$ represent the linguistic features produced by the last hidden state of BERT. By multiplying them, the cross-modal features $F = F_1 \times F_2$ are obtained. Then $F$ is linearly transformed to query, key and value by:
\begin{align}
q = W_q F \quad
k = W_k F \quad
v = W_v F 
\end{align}

\begin{figure}[htbp]
\centering
\subfigure[]
{
    \begin{minipage}[b]{.55\linewidth}
        \includegraphics[width=70mm]{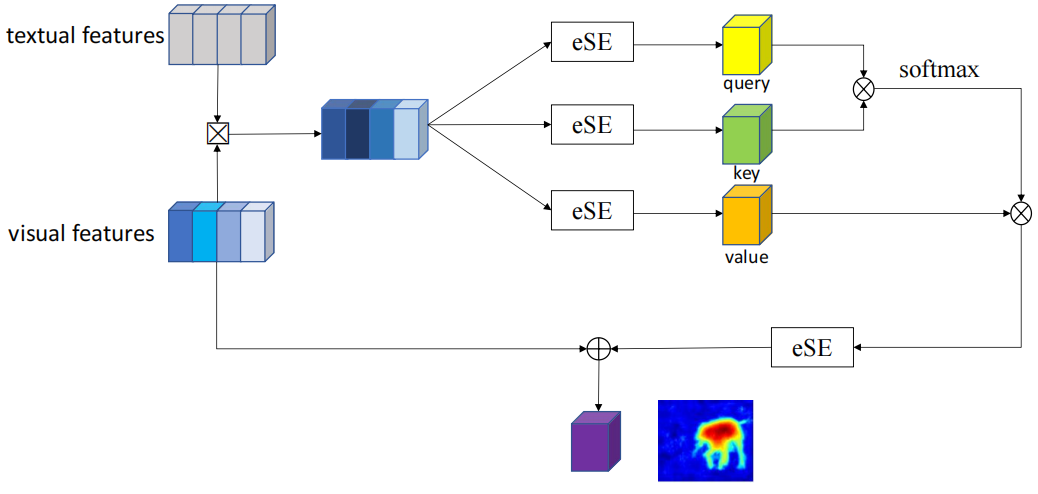}
    \end{minipage}%
}
\subfigure[]
{
 	\begin{minipage}[b]{.35\linewidth}
        \includegraphics[width=50mm]{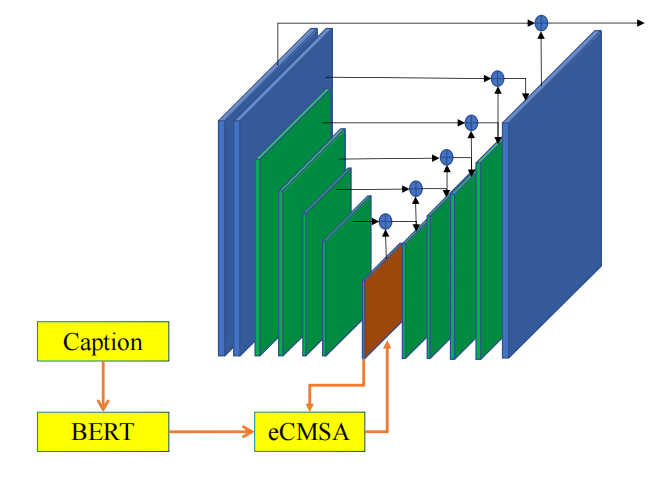}
    \end{minipage}%
}
\caption{(a): The structure of eCMSA. $\boxtimes$ indicates element-wise multiplication, $\otimes$ and $\oplus$ indicate matrix multiplication and element-wise summation. The eSE \cite{hu2018squeeze} module helps to add the channel attention to the eCMSA module. (b): The illustration of integrating the eCMSA module into the U-Net.}
\end{figure}

eSE is a module based on Squeeze-Excitation (SE) \cite{hu2018squeeze}. SE uses an average pooling layer to squeeze the spatial dependency, and extract the “more important” channels of feature maps with two fully connected (FC) layers followed by a sigmoid function. Unlike SE, eSE only uses one FC layer, so that there is no channel information loss. Introducing eSE module could help to add channel attentions to the original CMSA module and improve the representation ability.  The cross-modal self-attention is computed as:
\begin{equation}
\hat{V} = vSoftmax(q,k)    
\end{equation}

Then another channel attention transformation is used on $\hat{V}$ through an eSE module and we add the original visual features to $\hat{V}$ to get the final output features.

\subsection{U-Structure Networks with eCMSA}

There are various ways to combine U-Structure Networks like U-Net with the eCMSA module. The most direct and the simplest way is to add eCMSA at the bottom of the U-Net, because it costs less computational sources and is more likely to get higher performance. We take the feature maps $V\in R^{H\times W \times C}$ of the bottom layer of U-Net as the visual input of eCMSA, and add the output features of eCMSA $F_{eCMSA}$ to the original visual features to form a residual connection $V^{’}= V + F_{eCMSA}$, just like figure 4(b) shows.

Meanwhile, the same sort of combination is used in U$^2$Net and BASNet \cite{Qin_2019_CVPR}. U$^2$Net consists of many smaller U-Nets, the eCMSA can be added to the bottom of those smaller U-Nets. Also, eCMSA module can be added to the outer part of U$^2$Net, which is the whole U-structure network of U$^2$Net. For the BASNet, there are two parts, the Predict Module for predicting the coarse saliency maps and the Residual Refinement module for refining region and boundary drawbacks in the coarse saliency maps, both of which are in the U-structure shape. Then the eCMSA module can be added to the bottom of those two modules.

%\begin{figure}[!h]
%\centering\includegraphics[scale=0.65]{template/U2Net+eCMSA.PNG}
%\caption{The illustration of integrating the eCMSA module into the U$^2$Net.  }
%\label{fig:fire}
%\end{figure}

\section{Experiments}
We have conducted several experiments on the DUTS-CM, DUT-OMRON-CM and HKU-IS-CM datasets to prove the effect of the proposed method. All the experiments are completed on the GPU with a version of NVIDIA RTX A6000. The initial input size of images is 320*320, all batch sizes are set as 16, initial learning rates for all models are 5e-5, and the optimizer is AdamW \cite{loshchilov2017decoupled}. The datasets are augmented by horizontal flipping and random contrast. Also, all the images are randomly cropped from 320*320 to 288*288. And the cosine learning rate scheduler is applied in the training stage.

\subsection{Evaluation Metrics}
There are 5 metrics for evaluation: $F_\beta$ \cite{achanta2009frequency}, $E_m$ \cite{fan2018enhanced}, $MAE$\cite{liu2019simple}, $S_m$\cite{fan2017structure} and the infer time. 

(1)F-measure ($F_\beta$), is based on the PR curve. And $MaxF_\beta$ is the maximum $F_\beta$. This metric could evaluate both the precision and recall as:

\begin{equation}
F_\beta = \frac{(1+\beta^2)\times Precision\times Recall}{\beta^2 \times Precision\times Recall}    
\end{equation}

(2) E-measure ($E_m$) combines low-level and high-level matching information, which is effective and efficient. It can be computed as:

\begin{equation}
E_m = \frac{1}{ w \times h}\sum_{r=1}^{H} \sum_{c=1}^{W}\phi_ {FM}\{r,c\}    
\end{equation}

$\phi_{FM}$ is the enhanced alignment matrix.

(3)MAE, means the Mean Absolute Error, which measures the pixel-level difference between the predicted mask image and the ground truth mask image. %$MAE$ is computed as: 

%\begin{equation}
%MAE = \frac{1}{H\times W}\sum_{r=1}^{H} \sum_{c=1}^{W}|P(r,c) - G(r,c)|   
%\end{equation}

%P means the continuous prediction map and G means the ground truth map.

(4)S-measure ($S_m$), which means structure-measure. It measures both the region-aware and the object-aware structural similarity.

%\begin{equation}
%S_m = \alpha \times S_o + (1-\alpha)\times S_r
%\end{equation}

(5) Infer time is the abbreviation of inference time. it indicates how much time is needed to process a single image.

\subsection{Results and Discussions}
In the experiments, the baseline models are U-Net, U$^2$Net and BASNet models. Table 1 shows the effect of adding the eCMSA module to the original CNN based models. For instance, compared with U-Net, U-Net+eCMSA is fed with both visual and linguistic information, which is also the best model achieved in experiments related to U-Net. It can get higher performance on all the metrics, $F\_beta$, $E\_beta$ ,$MAE$ and $Smeasure$. Figure 5 gives us an intuitive impression of the performance of these six models. But with improvement of the performance, the size of the model also increases a lot. 

\begin{table*}[]
    \renewcommand\arraystretch{1}
    \centering
    \vspace{1mm}
    \scriptsize
    \caption{The experiments of U-Net, U$^2$Net and BASNet on DUTS-TE. eCMSA indicates the eCMSA module has been applied in the model. }
    \begin{tabular}{l|ccccc}
    \hline
     Method & Max$F_\beta\uparrow$  & MAE$\downarrow$ & Max$E_m$ $\uparrow$ & $S_m\uparrow$ & infer time/s $\downarrow$
        \\ \hline
        U-Net & 0.805 & 0.060 & 0.888 & 0.843 & 0.005
        \\ 
        U-Net+eCMSA & 0.825 & 0.053 & 0.899 & 0.855 &0.022
        \\ 
        U$^2$Net & 0.846 & 0.045 & 0.911 &0.871 &0.024  
        \\ 
        U$^2$Net+eCMSA & 0.860 & 0.041 & 0.918 &0.880 &0.031   
        \\ 
        BASNet & 0.862 & 0.042 & 0.920 &0.881 &0.011
        \\ 
        BASNet+eCMSA & 0.868 & 0.039 & 0.922 &0.885 &0.021
        \\ \hline
        
    \end{tabular}
\vspace{-5mm}
\end{table*}

And we could see similar things happened on the U$^2$Net and BASNet. Just like U-Net+eCMSA, U$^2$Net+eCMSA and BASNet+eCMSA are the best models achieved in the experiments, based on their prototypes. After adding eCMSA modules, their performances on DUTS-TE are improved, but with the cost of higher inference time. 

From table 2, we compare our methods with different classical methods like PoolNet and MINet. The result shows that our methods could perform better than most of these methods rendered in the table, proving the competitiveness of our methods. One more thing worth noting is that on the DUTS-CM dataset, the BASNet can have better performance than U$^2$Net and U-Net, and U$^2$Net has better performance than U-Net. But in table 3 we could see the BASNet could not outperform U$^2$Net on the DUT-OMRON dataset. And when the eCMSA module is joined to BASNet, the new model BASNet$|$prednet could get a better performance than the original BASNet model on DUTS-TE, but the improvement on the other two datasets is not obvious. One probable reason for this is that BASNet lacks a good generalization ability, so it could perform well on a specific dataset but could not copy this performance to other datasets.

\begin{table*}[ht]
    \renewcommand\arraystretch{1}
    \centering
    \scriptsize
    \setlength\tabcolsep{1pt}
    \vspace{1mm}
    \caption{Comparison with the state-of-the-art methods. }
    \begin{tabular}{lccccccccccccccccccc}
    \hline
        \multirow{2}{*}{Method}&
        \multicolumn{4}{c}{DUTS-TE}&
        \multicolumn{4}{c}{DUT-OMRON-CM}&
        \multicolumn{4}{c}{HKU-IS-CM}
     \\
     \cline{2-13}
         & Max$F_\beta$ & MAE & Max$E_m$ & $S_m$
         & Max$F_\beta$ & MAE & Max$E_m$ & $S_m$
         & Max$F_\beta$ & MAE & Max$E_m$ & $S_m$ \\
        \hline
        UCF \cite{zhang2017learning} 
        & 0.742 & 0.112 & 0.843 & 0.782
        & 0.698 & 0.120 & 0.821 &0.760 
        & 0.874 & 0.062 & 0.926 &0.875 \\ 
        BiMCFEM \cite{zhang2018bi} 
        & 0.828 & 0.049 & 0.907 &0.862
        & 0.734 & 0.064 & 0.848 &0.809
        & 0.910 & 0.039 & 0.950 &0.907 \\ 
        Amulet \cite{zhang2017amulet} 
        & 0.750 & 0.085 & 0.851 & 0.804
        & 0.715 & 0.098 & 0.793 &0.781
        & 0.887 & 0.051 & 0.933 &0.886 \\ 
        AFNet \cite{feng2019attentive} 
        & 0.839 & 0.046 & 0.910 & 0.867
        & 0.759 & 0.057 & 0.861 &0.826
        & 0.910 & 0.036 & 0.949 &0.905 \\ 
        PiCANet \cite{liu2020picanet} 
        & 0.827 & 0.054 & 0.907 & 0.861
        & 0.759 & 0.068 & 0.866 &0.826
        & 0.908 & 0.042 & 0.949 &0.906 \\ 
        CPD \cite{wu2019cascaded} 
        & 0.839 & 0.043 & 0.912 & 0.867 
        & 0.747 & 0.057 & 0.856 &0.818
        & 0.909 & 0.033 & 0.948 &0.904 \\ 
        SCWSSOD \cite{yu2021structure} 
        & 0.823 & 0.049 & 0.907 & 0.841
        & 0.756 & 0.060 & 0.868 & 0.812
        & 0.898 & 0.038 & 0.943 &0.882 \\ 
        TSPOANet \cite{liu2019employing} 
        & 0.829 & 0.049 & 0.907 & 0.860
        & 0.750 & 0.052 & 0.858 &0.818
        & 0.909 & 0.038 & 0.950 &0.902 \\ 
        PoolNet \cite{liu2019simple} 
        & 0.853 & 0.042 & 0.918 & 0.879
        & 0.769 & 0.056 & 0.869 &0.832
        & 0.920 & 0.032 & 0.955 &0.916 \\ 
        GateNet \cite{zhao2020suppress} 
        & 0.847 & 0.045 & 0.915 & 0.870
        & 0.754 & 0.061 & 0.858 &0.821
        & 0.916 & 0.036 & 0.951 &0.910 \\ 
        MINet \cite{pang2020multi} 
        & 0.852 & 0.039 & 0.917 & 0.875
        & 0.751 & 0.057 & 0.856 &0.822
        & 0.918 & 0.031 & 0.952 &0.912 \\ 
        U$^2$Net \cite{qin2020u2} 
        & 0.846 & 0.045 & 0.911 &0.871
        & 0.791 & 0.054 & 0.882 &0.845
        & 0.921 & 0.032 & 0.952 &0.913\\ 
        BASNet \cite{Qin_2019_CVPR} 
        & 0.862 & 0.042 & 0.920 &0.881
        & 0.783 & 0.057 & 0.875 &0.840
        & 0.923 & 0.032 & 0.953 &0.914\\ 
        %VST-RGB \cite{liu2021visual} 
        %& 0.878 & 0.037 & 0.939 &0.896
        %& 0.800 & 0.058 & 0.888 &0.850
        %& 0.937 & 0.029 & 0.968 &0.928\\
        U$^2$Net+eCMSA 
        & 0.860 & 0.041 & 0.918 &0.880
        & 0.802 & 0.049 & 0.889 &0.853
        & 0.925 & 0.031 & 0.955 &0.916 \\ 
        BASNet+eCSMA 
        & 0.868 & 0.039 & 0.922 &0.885
        & 0.781 & 0.053 & 0.877 &0.841
        & 0.924 & 0.033 & 0.952 &0.914\\ \hline
        %VST-RGB+eCMSA & 0.855 & 0.880 & 0.035 & 0.924 & 0.938 &0.895& 
        %0.783 & 0.802 & 0.053 & 0.876 & 0.888 &0.852& 
        %0.913 & 0.936 & 0.028 & 0.953 & 0.965 &0.926\\ \hline
    \end{tabular}
\end{table*}

\begin{figure*}[!h]
\centering\includegraphics[width=110mm]{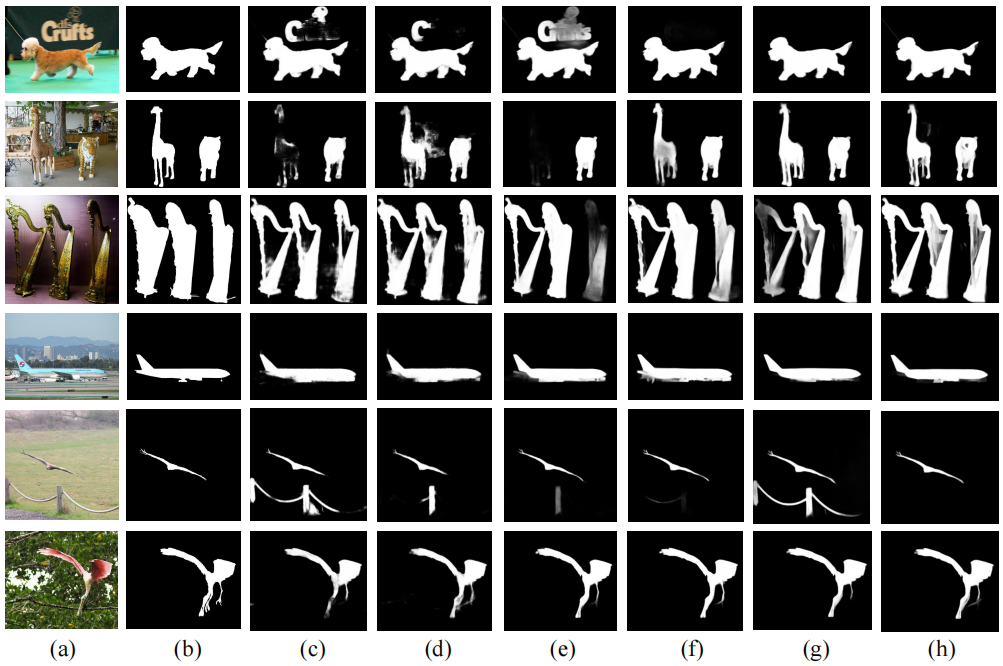}
\caption{The Qualitative results on different models:(a) image, (b) ground truth (c) U-Net, (d) U-Net+eCMSA, (e) U$^2$Net, (f) U$^2$Net+eCMSA, (g) BASNet, (h) BASNet+eCMSA  }
\end{figure*}

In the next part, the ablation study will show the specific effects of different modules and the positions of connecting eCMSA modules with the original models. Also, it will study the influence of different types of words in the sentences.

\subsection{Ablation Study}
\textbf{The effectiveness of eSE module and residual connection}
From table 2, we can see that when the eCMSA module is added to the bottom of the U-Net, we have U-Net$|$1. And it achieved higher performance on all three datasets. And after removing the eSE module or the residual connection from the eCMSA module, we could see there is a decrease in performance in the three datasets, which indicates the effectiveness of the channel attention provided by eSE modules and the effectiveness of residual connection.

\textbf{Ablation study of eCMSA modules}
The eCMSA modules can be connected to the different positions of the three models. Then we can see a trend that the larger the model, the better the performance can be achieved. Adding the eCMSA modules to the parts which are closer to the bottom of the network is more likely to get higher performance. Just like U$^2$Net$|$in:1 and U$^2$Net$|$in:2 perform better than U$^2$Net$|$in:3 and U$^2$Net$|$in:4. The probable cause of it is that the embedding vectors generated from the sentences fit better with the visual embedding information, which represents a higher representation, because the textual embedding is also a kind of higher representation of linguistic information. Also, we could see that BASNet$|$prednet performs better than BASNet$|$refunet on DUTS and DUT-OMRON datasets, and there is no significant difference between their performance on HKU-IS. This may help to further prove the hypothesis aforementioned, because the visual features from the bottom layer of the residual refinement module can be taken as a lower representation compared with the visual features from the bottom layer of the predict module of BASNet. 

In addition, adding multiple eCMSA modules to different parts of the U-structure networks may help to produce better performance than adding a single eCMSA module to the networks, but always with the cost of higher inference time. U-Net$|$1-2 means the eCMSA modules are added to the first and second layers of U-Net (from bottom up), and we can roughly say that U-Net$|$1-2 performs better than both U-Net$|$1 and U-Net$|$2. And generally, U$^2$Net$|$in:1-2, U$^2$Net$|$in:1-3 and U$^2$Net$|$in:1-4 perform better than U$^2$Net$|$in:1, U$^2$Net$|$in:2, U$^2$Net$|$in:3, and U$^2$Net$|$in:4. And adding the eCMSA module to both the inner and outer layers may further improve the performance, just as U$^2$Net$|$in:1-3$|$out:1 shows.

\begin{table*}[]
    \renewcommand\arraystretch{1}
    \centering
    \scriptsize
    \setlength\tabcolsep{1pt}
    \caption{ For unet$|$i, i means the eCMSA module is added at the i-th layer(from bottom up) of the U-shape network. For U$^2$Net$|$in and U$^2$Net$|$out, in and out indicate the inner and outer layers of the U$^2$Net. no-eSE and no-res indicate that eSE modules and residual connections are removed, no-color and no-objects indicate that the words related to color or name of objects are removed. And BASNet$|$refunet and BASNet$|$prednet indicate that the eCMSA module is added to the residual refinement module and the predict module of BASNet. }
    \begin{tabular}{lcccccccccccccccccccc}
    \hline
    \multirow{2}{*}{Method}&
    \multicolumn{4}{c}{DUTS-TE}&
    \multicolumn{4}{c}{DUT-OMRON-CM}&
    \multicolumn{4}{c}{HKU-IS-CM}
     \\
    \cline{2-14}
         & Max$F_\beta$ & MAE & Max$E_\beta$ & $S_m$
         & Max$F_\beta$ & MAE & Max$E_\beta$ & $S_m$
         & Max$F_\beta$ & MAE & Max$E_\beta$ & $S_m$ & infer time/s\\
        \hline
        U-Net 
        & 0.805 & 0.060 & 0.888 & 0.843
        & 0.767 & 0.068 & 0.868 & 0.824
        & 0.906 & 0.039 & 0.944 &0.901 &0.005\\ 
        U-Net\_1 
        & 0.820 & 0.055 & 0.898 & 0.853
        & 0.780 & 0.062 & 0.880 &0.832
        & 0.910 & 0.038 & 0.946 &0.903 &0.015\\
        U-Net$|$2 
        & 0.820 & 0.054 & 0.896 & 0.851
        & 0.780 & 0.060 & 0.879 & 0.833
        & 0.906 & 0.041 & 0.941 &0.897 &0.020\\ 
        U-Net$|$1-2 
        & 0.825 & 0.053 & 0.899 & 0.855
        & 0.784 & 0.060 & 0.882 &0.835
        & 0.909 & 0.039 & 0.944 &0.901 &0.022\\ 
        U-Net$|$1$|$no-eSE
        & 0.807 & 0.060 & 0.887 & 0.843
        & 0.766 & 0.068 & 0.869 &0.822
        & 0.909 & 0.039 & 0.945 &0.902 &0.015\\ 
        U-Net$|$1$|$no-res 
        & 0.808 & 0.070 & 0.886 & 0.844
        & 0.765 & 0.070 & 0.867 &0.820
        & 0.904 & 0.041 & 0.942 &0.898 &0.015\\ 
        U-Net$|$1$|$no-color 
        & 0.819 & 0.055 & 0.897 & 0.852
        & 0.780 & 0.062 & 0.880 &0.832
        & 0.910 & 0.038 & 0.946 &0.903 &0.015\\ 
        U-Net$|$1$|$no-objects 
        & 0.807 & 0.060 & 0.889 & 0.842
        & 0.771 & 0.065 & 0.872 &0.826
        & 0.907 & 0.040 & 0.943 &0.900 &0.015\\ \hline
        U$^2$Net 
        & 0.846 & 0.045 & 0.911 &0.871
        & 0.791 & 0.054 & 0.882 &0.845
        & 0.921 & 0.032 & 0.952 &0.913 &0.020\\ 
        U$^2$Net$|$in:1 
        & 0.853 & 0.043 & 0.913 & 0.877
        & 0.794 & 0.051 & 0.883 & 0.850
        & 0.922 & 0.032 & 0.952 &0.913 &0.029\\ 
        U$^2$Net$|$in:2 
        & 0.851 & 0.043 & 0.911 &0.874
        & 0.790 & 0.050 & 0.882 &0.847
        & 0.918 & 0.033 & 0.948 &0.909 &0.028\\
        U$^2$Net$|$in:3 
        & 0.847 & 0.046 & 0.908 & 0.870
        & 0.787 & 0.054 & 0.879 & 0.843
        &0.919 & 0.033  & 0.951 & 0.911 &0.028\\
        U$^2$Net$|$in:4 
        & 0.841 & 0.048 & 0.905 &0.866
        & 0.790 & 0.052 & 0.884 &0.847
        &0.915 & 0.035  & 0.947 &0.907 & 0.028\\
        U$^2$Net$|$in:1-2 
        & 0.855 & 0.042 & 0.917 &0.878
        & 0.799 & 0.050 & 0.889 &0.853
        & 0.920 & 0.033 & 0.950 &0.911 & 0.030\\
        U$^2$Net$|$in:1-3 
        & 0.859 & 0.042 & 0.918 &0.879
        & 0.793 & 0.052 & 0.882 &0.848
        & 0.923 & 0.032 & 0.953 &0.914 & 0.032\\
        U$^2$Net$|$in:1-4 
        & 0.857 & 0.043 & 0.917 &0.878
        & 0.801 & 0.051 & 0.893 &0.854
        & 0.925 & 0.032 & 0.954 &0.915 & 0.034\\
        U$^2$Net$|$out:1 
        & 0.844 & 0.047 & 0.907 &0.870
        & 0.791 & 0.055 & 0.879 &0.846
        & 0.921 & 0.033 & 0.952 &0.913 & 0.027\\
        U$^2$Net$|$in:1-3$|$out:1 
        & 0.860 & 0.041 & 0.918 & 0.880
        & 0.802 & 0.049 & 0.889 & 0.853
        & 0.925 & 0.031 & 0.955 & 0.916 &0.033\\
        \hline
        BASNet 
        & 0.862 & 0.042 & 0.920 &0.881
        & 0.783 & 0.057 & 0.875 &0.840
        & 0.923 & 0.032 & 0.953 &0.914 & 0.013\\
        BASNet$|$refunet 
        & 0.863 & 0.040 & 0.919 &0.882
        & 0.778 & 0.054 & 0.872 &0.838
        & 0.923 & 0.031 & 0.953 &0.913 & 0.021\\
        BASNet$|$prednet 
        & 0.868 & 0.039 
        & 0.922 & 0.885
        & 0.781 & 0.053 & 0.877 & 0.841
        & 0.924 & 0.033 & 0.952 &0.914 &0.021\\ \hline
        %BASNet\_F\&refunet & \textbf{0.847} & \textbf{0.869} & \textbf{0.039} & \textbf{0910} & \textbf{0.924} &\textbf{0.886}
        %& 0.766 & 0.781 & 0.052 & 0.859 & 0.878 &0.841
        %& 0.905 & 0.924 & 0.031 & 0.942 & 0.953 &0.915&0.021\\ \hline
        %BASNet\_F\&refunet & \textbf{0.847} & \textbf{0.869} & \textbf{0.039} & \textbf{0910} & \textbf{0.924} &\textbf{0.886}
        %& 0.774 & 0.789 & 0.054 & 0.866 & 0.882 &0.844
        %& 0.905 & 0.924 & 0.031 & 0.942 & 0.953 &0.915&0.021\\ \hline
    \end{tabular}
\end{table*}

\textbf{Different types of captions}
For the captions helping to describe the objects in the images, there are mainly two types of words in each caption: words related to the names of objects and words related to colors. In the experiment, we separately removed the words related to names of objects (entities) or colors by replacing the specific words with ‘UNK’ (unknown) and tested how the model would perform without the existence of these two kinds of words. Based on the model U-Net$|$1 in table 3, we can see that after removing the entity related and color related words, both the performances of the model got worse. This proves the effectiveness of depicting entities and colors in the captions. And removing the entities related words will lead to lower performance than removing the color related words, which indicates that the network collects more information from the entities rather than colors.

\section{Conclusion}
In this work, we propose a new dataset called DUTS-CM by adding the linguistic labels to the original DUTS dataset, and labeled the DUT-OMRON and HKU-IS dataset based on the human labeled DUTS-CM dataset. The DUTS-CM could also help to provide a platform for other researchers to conduct their own multimodal study on the salient object detection task. Also, we propose the eCMSA module to integrate the textual and visual information effectively and achieved competitive performance on the three public datasets. We also conduct several ablation studies to explore some problems, including the effectiveness of eSE modules, residual connection and different types of words in captions, and the effectiveness of positions adding eCMSA modules to a U-structure network. Some future work, such as replacing the convolutional layers with the self-attention layers (with the structure similar to eCMSA) could be done to further improve the performance.

\bibliography{egbib}
\end{document}